\pdfoutput=1
\PassOptionsToPackage{dvipsnames}{xcolor}
\documentclass[11pt]{article}

\usepackage{ACL2023}

\usepackage{times}
\usepackage{latexsym}

\usepackage[T1]{fontenc}

\usepackage[utf8]{inputenc}

\usepackage{microtype}

\usepackage{inconsolata}

\usepackage{graphicx}
\usepackage{enumitem}

\usepackage{pifont}
\newcommand{\cmark}{\ding{51}}%
\newcommand{\xmark}{\ding{55}}%

\usepackage{amsfonts}
\usepackage{multirow}
\usepackage{booktabs}
\usepackage{amsmath}
\usepackage{enumerate}
\usepackage{wrapfig}
\usepackage{todonotes}
\usepackage{makecell}
\usepackage{float}
\usepackage{array}
\usepackage{pdfpages}
\usepackage{algorithm}
\usepackage{graphicx}
\usepackage{algpseudocode}
\usepackage{algorithmicx}
\usepackage[dvipsnames]{xcolor}

\usepackage{color, colortbl}
\definecolor{LRed}{rgb}{1,.9,.9}
\definecolor{LGreen}{rgb}{.9,1,0.9}
\definecolor{LBlue}{rgb}{.9,.9,1}
\definecolor{LYellow}{rgb}{1,1,0.9}

\definecolor{lightblue}{rgb}{0.68, 0.85, 0.9}
\definecolor{lavender}{rgb}{0.9, 0.9, 0.98}
\definecolor{lightyellow}{rgb}{1.0, 1.0, 0.88}
\definecolor{magicmint}{rgb}{0.67, 0.94, 0.82}
\definecolor{palepink}{rgb}{0.98, 0.85, 0.87}
\definecolor{bubbles}{rgb}{0.91, 1.0, 1.0}

\title{UniS-MMC: Multimodal Classification via Unimodality-supervised Multimodal Contrastive Learning}

\author{Heqing Zou, Meng Shen, Chen Chen, Yuchen Hu, Deepu Rajan, Eng Siong Chng \\
    Nanyang Technological Universityy, Singapore \\
   \small \texttt{\{heqing001, meng005, chen1436, yuchen005\}@e.ntu.edu.sg, \{asdrajan, aseschng\}ntu.edu.sg}\\
}

\begin{document}
\maketitle
\begin{abstract}
Multimodal learning aims to imitate human beings to acquire complementary information from multiple modalities for various downstream tasks. However, traditional aggregation-based multimodal fusion methods ignore the inter-modality relationship, treat each modality equally, suffer sensor noise, and thus reduce multimodal learning performance. In this work, we propose a novel multimodal contrastive method to explore more reliable multimodal representations under the weak supervision of unimodal predicting.
Specifically, we first capture task-related unimodal representations and the unimodal predictions from the introduced unimodal predicting task. Then the unimodal representations are aligned with the more effective one by the designed multimodal contrastive method under the supervision of the unimodal predictions.
Experimental results with fused features on two image-text classification benchmarks UPMC-Food-101 and N24News show that our proposed \textbf{Uni}modality-\textbf{S}upervised \textbf{M}ulti\textbf{M}odal \textbf{C}ontrastive (\textbf{UniS-MMC}) learning method outperforms current state-of-the-art multimodal methods. The detailed ablation study and analysis further demonstrate the advantage of our proposed method. 
\end{abstract}

\begin{figure}[t]
    \centering
    \includegraphics[width=1\columnwidth]{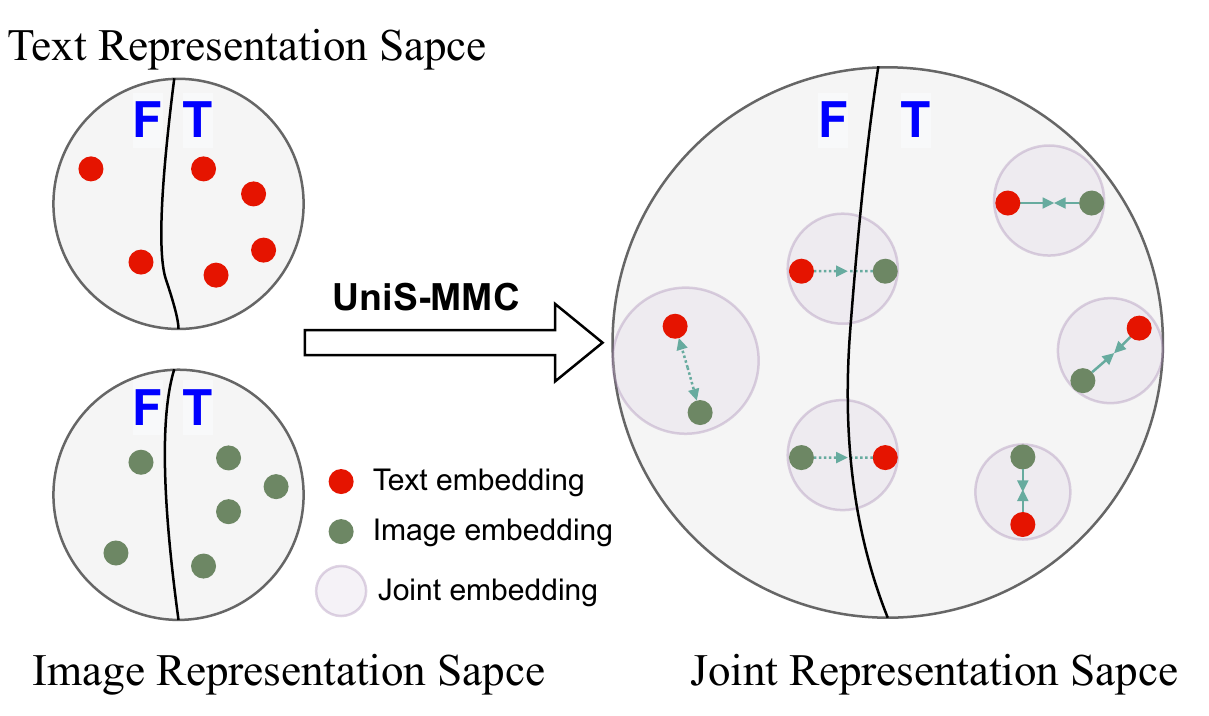}
    \caption{ Unimodal representation of a single modality can be either effective or not. The effectiveness of different unimodal representations from the same sample also varies. To empower the interaction between modalities, our proposed method aligns the unimodal representation to the effective modality sample-wise and makes full use of the effective unimodal representation under the supervision of the unimodal prediction (F and T represent correct and incorrect predictions, respectively).}
    \label{fig01}
\end{figure}

\section{Introduction}

Social media has emerged as an important avenue for communication. The content is often multimodal, e.g., via text, speech, audio, and videos. Multimodal tasks that employ multiple data sources include image-text classification and emotion recognition, which could be used for specific applications in daily life, such as web search \citep{chang2022webqa}, guide robot \citep{moon2019observation}. Hence, there is a need for an effective representation strategy for multimodal content. A common way is to fuse unimodal representations. 
Despite the recent progress in obtaining effective unimodal representations from large pre-trained models \citep{devlin-etal-2019-bert, liu2019roberta, dosovitskiy2021an}, fusing for developing more trustworthy and complementary multimodal representations remains a challenging problem in the multimodal learning area.

To solve the multimodal fusion problem, researchers propose aggregation-based fusion methods to combine unimodal representations. These methods include aggregating unimodal features \citep{castellano2008emotion, nagrani2021attention}, aggregating unimodal decisions \citep{ramirez2011modeling, tian2020uno}, and aggregating both \citep{wu2022characterizing} of them. 
However, these aggregation-based methods ignore the relation between modalities that affects the performance of multimodal tasks \cite{udandarao2020cobra}.
To solve this issue, the alignment-based fusion methods are introduced to strengthen the inter-modality relationship by aligning the embeddings among different modalities. Existing alignment-based methods can be divided into two categories: architecture-based and contrastive-based. The architecture-based methods introduce a specific module for mapping features to the same space\citep{wang2016learning} or design an adaption module before minimizing the spatial distance between source and auxiliary modal distributions \citep{song2020modality}. On the other hand, the contrastive--based methods efficiently align different modality representations through the contrastive learning on paired modalities \cite{liu2021contrastive,zolfaghari2021crossclr,mai2022hybrid}.

The unsupervised multimodal contrastive methods directly regard the modality pairs from the same samples as positive pairs and those modality pairs from different samples as negative pairs to pull together the unimodal representations of paired modalities and pull apart the unimodal representations of unpaired modalities in the embedding space. \citep{tian2020contrastive, akbari2021vatt, zolfaghari2021crossclr, liu2021contrastive, zhang2021cross, taleb2022contig}. Supervised multimodal contrastive methods are proposed to treat sample pairs with the same label as positive pairs and sample pairs with a different label as negative pairs in the mini-batch \citep{zhang2021supervised, pinitas2022supervised}. In this way, the unimodal representations with the same semantics will be clustered.

Despite their effectiveness in learning the correspondence among modalities, these contrastive-based multimodal learning methods still meet with problems with the sensor noise in the in-the-wild datasets \citep{mittal2020m3er}. The current methods always treat each modality equally and ignore the difference of the role for different modalities,  The final decisions will be negatively affected by those samples with inefficient unimodal representations and thus can not provide trustworthy multimodal representations. In this work, we aim to learn trustworthy multimodal representations by aligning unimodal representations towards the effective modality, considering modality effectiveness in addition to strengthening relationships between modalities. The modality effectiveness is decided by the unimodal prediction and the contrastive learning is under the weak supervision information from the unimodal prediction. As shown in Figure \ref{fig01}, the unimodal representations will be aligned towards those with correct unimodal predictions. In summary, our contributions are:

\begin{itemize}[leftmargin=*]
\setlength\itemsep{0em}
    \item  To facilitate the inter-modality relationship for multimodal classification, we combine the aggregation-based and alignment-based fusion methods to create a joint representation.  
    \item  We propose UniS-MMC to efficiently align the representation to the effective modality under weak supervision of unimodal prediction to address the issue of different contributions from the modailities.
    \item  Extensive experiments on two image-text classification benchmarks, UPMC-Food-101 \citep{wang2015recipe} and N24News \citep{wang-etal-2022-n24news} demonstrate the effectiveness of our proposed method.
\end{itemize}

\section{Related Work}
In this section, we will introduce the related work on contrastive learning and multimodal learning.

\subsection{Contrastive Learning}

Contrastive learning \citep{hadsell2006dimensionality, oord2018representation, qin-joty-2022-continual} captures distinguishable representations by drawing positive pairs closer and pushing negative pairs farther contrastively. In addition to the above single-modality representation learning, contrastive methods for multiple modalities are also widely explored. The common methods \citep{radford2021learning, jia2021scaling, kamath2021mdetr, li2021align, zhang2022mcse, taleb2022contig, chen2022interactive} leverage the cross-modal contrastive matching to align two different modalities and learn the inter-modality correspondence. Except the inter-modality contrastive, Visual-Semantic Contrastive \citep{yuan2021multimodal}, XMC-GAN \citep{zhang2021cross} and CrossPoint \citep{afham2022crosspoint} also introduce the intra-modality contrastive for representation learning. Besides, CrossCLR \citep{zolfaghari2021crossclr} removes the highly related samples from negative samples to avoid the bias of false negatives. GMC \citep{Poklukar2022GeometricMC} builds the contrastive learning process between the modality-specific representations and the global representations of all modalities instead of the cross-modal representations.

\begin{figure*}[ht]
    \centering
    \includegraphics[width=1\textwidth]{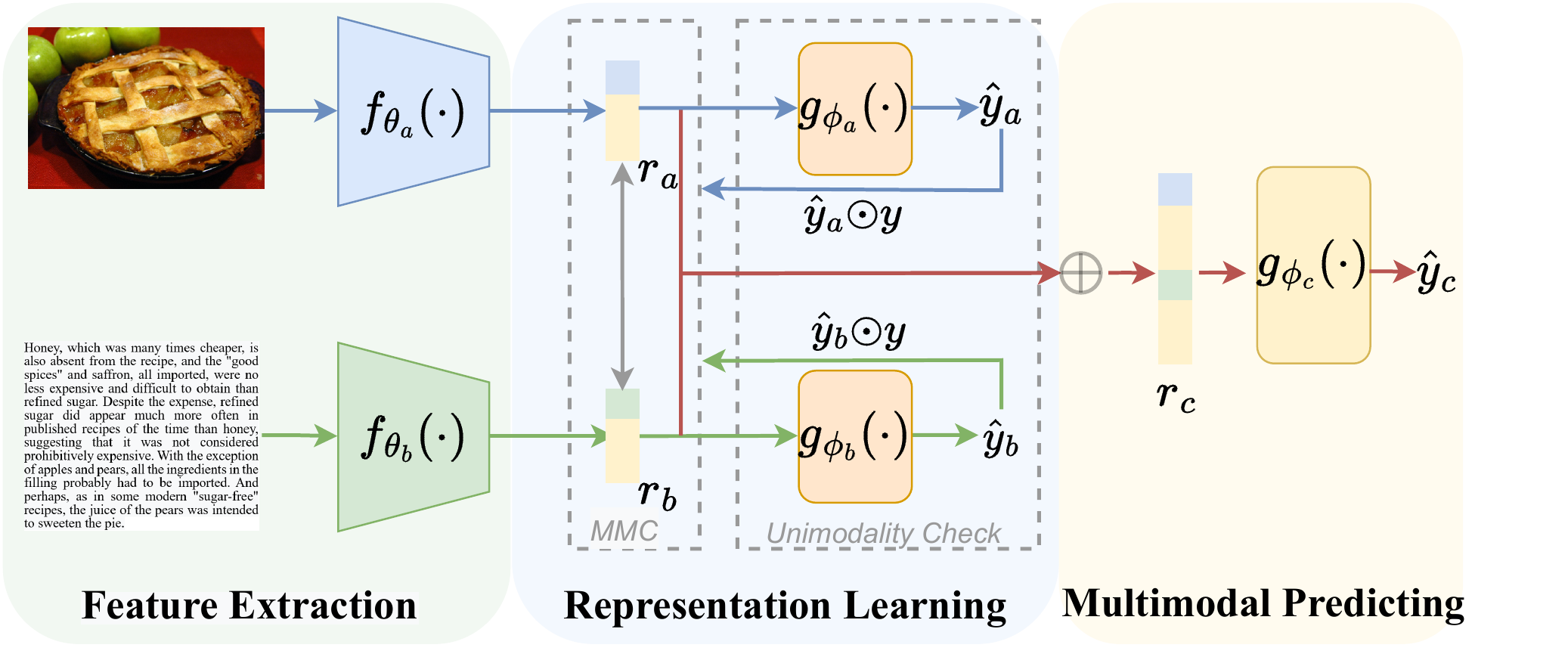}
    \caption{The framework for our proposed UniS-MMC.}
    \label{fig02}
\end{figure*}

\subsection{Multimodal Learning}

Multimodal learning is expected to build models based on multiple modalities and to improve the general performance from the joint representation \citep{ngiam2011multimodal,baltruvsaitis2018multimodal,gao2020survey}. The fusion operation among multiple modalities is one of the key topics in multimodal learning to help the modalities complement each other \citep{wang2021survey}. Multimodal fusion methods are generally categorized into two types: alignment-based and aggregation-based fusion \citep{baltruvsaitis2018multimodal}. Alignment-based fusion \citep{gretton2012kernel, song2020modality} aligns multimodal features by increasing the modal similarity to capture the modality-invariant features. Aggregation-based methods choose to create the joint multimodal representations by combining the participating unimodal features (early-fusion, \citet{kalfaoglu2020late,nagrani2021attention,zou2022speech}), unimodal decisions (late-fusion, \citet{tian2020uno, huang2022modality}) and both (hybrid-fusion, \citet{wu2022characterizing}). In addition to these joint-representation generating methods, some works further propose to evaluate the attended modalities and features before fusing. M3ER \citep{mittal2020m3er} conducts a modality check step to finding those modalities with small correlation and Multimodal Dynamics \citep{Han_2022_CVPR} evaluates both the feature- and modality-level informativeness during extracting unimodal representations.

\section{Methodology}
In this section, we present our method called UniS-MMC for multimodal fusion.

\subsection{Notation}

Suppose we have the training data set $\mathcal{D} = \{ \{x^n_m\}_{m=1}^M, y^n\}_{n=1}^N$ that  contains $\mathit{N}$ samples $\mathcal{X} = \{x^n_m\in{\mathbb{R}^{d_m}}\}_{m=1}^M$ of $\mathit{M}$ modalities and $\mathit{N}$ corresponding labels $\mathcal{Y} = \{ y^n\}_{n=1}^N$ from $\mathit{K}$ categories.  As shown in Figure \ref{fig02}, the unimodal representations of modality $a$ and $b$ are extracted from the respective encoders ${f_{\theta}}_a$ and ${f_{\theta}}_b$. Following the parameter sharing method in the multi-task learning \citep{DBLP:conf/iclr/PilaultEP21, Bhattacharjee_2022_CVPR}, the representations are shared directly between unimodal prediction tasks and the following multimodal prediction task. With weak supervision information produced from the respective unimodal classifier ${g_{\phi}}_a$ and ${g_{\phi}}_b$, the final prediction is finally learned based on the updated multimodal representations $r_c$ and the multimodal classifier ${g_{\phi}}_c$.

\begin{figure}[t]
    \centering    \includegraphics[width=1\columnwidth]{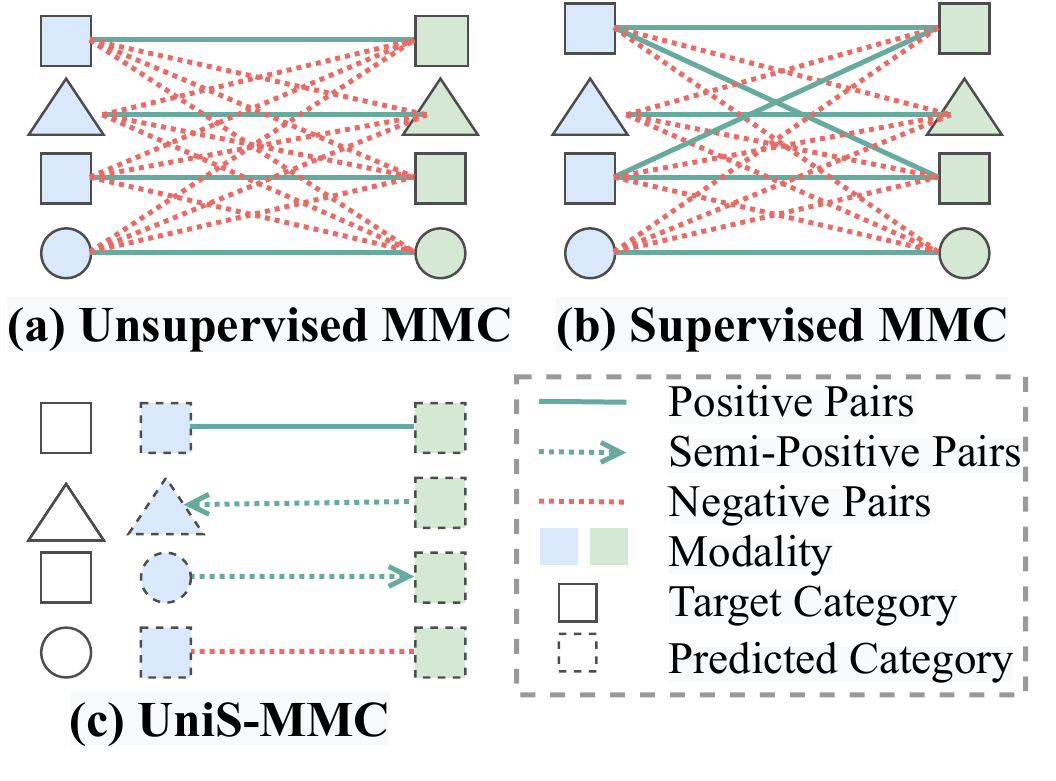}
    \caption{The relationship comparison between two modalities in training mini-batch of (a) unsupervised MMC, (b) supervised MMC and (c) UniS-MMC.}
    \label{fig03}
\end{figure}

\subsection{Unimodality-supervised Multimodal Contrastive Learning}

First, the unimodal representations are extracted from the raw data of each modality by the pretrained encoders. We introduce the uni-modality check step to generate the weak supervision for checking the effectiveness of each unimodal representation. Then we illustrate how we design the unimodality-supervised multimodal contrastive learning method among multiple modalities to learn the multimodal representations.

\subsubsection{Modality Encoder}

Given multimodal training data $\{\mathbf{x}_m\}_{m=1}^M$, the raw unimodal data of modality $m$ are firstly processed with respective encoders to obtain the hidden representations.
We denote the learned hidden representation $f_{\theta_m}(\mathbf{x}_m)$ of modality $m$ as $\mathbf{r}_m$. We use the pretrained ViT \cite{dosovitskiy2021an} as the feature encoder for images in both UPMC Food-101 and N24News datasets. We use only the pretrained BERT \citep{devlin-etal-2019-bert} as the feature encoder for the textual description in these datasets. Besides, we also try the pretrained  RoBERTa \cite{liu2019roberta} for text sources in N24News.

\subsubsection{Unimodality Check}

\textbf{Unimodal prediction.} Different from the common aggregation-based multimodal learning methods which only use the unimodal learned representations for fusion, our method also use the unimodal representations as inputs to the unimodal predicting tasks. The classification module can be regarded as a probabilistic model: $g_{\phi}: \mathcal{R} \rightarrow \mathcal{P}$, which maps the hidden representation to a predictive distribution $\mathbf{p}(\mathbf{y}\, |\, \mathbf{r})$. For a unimodal predicting task, the predictive distribution is only based on the output of the unimodal classifier. The learning objective of the unimodal predicting task is to minimize each unimodal prediction loss:

\begin{equation} \small
    \begin{aligned}
        \mathcal{L}_{uni}  = -\sum_{m=1}^M\sum_{k=1}^K{y^k\log{p_m^k}},
        \label{eq:eq04}
    \end{aligned}
\end{equation}
where $y^{k}$ is the $k$-th element category label and $[p_m^1;p_m^2;...;p_m^K] = \mathbf{p}_m(\mathbf{y}\, |\, \mathbf{r}_m)$ is the softmax output of unimodal classifiers on modality $m$.

\textbf{Unimodality effectiveness.} The above unimodal prediction results are used to check the supervised information for deciding the effectiveness of each modality. The unimodal representation with correct prediction is regarded as the effective representation for providing the information to the target label. Alternately, the unimodal representation with the wrong prediction is regarded as an ineffective representation. 

\begin{table}[h]
\caption{Contrastive settings.}
\vspace{-1em}
\setlength\tabcolsep{3pt}
\label{tab00}
\begin{center}
\scalebox{0.86}{\begin{tabular}{c|cccccc}
\toprule
    \text{Uni-Prediction} &  & \text{Modality} $a$ & &  \text{Modality} $b$ & &  \text{Category}  \\
    \midrule
    0 &  & True & & True &  & Positive  \\ 
    1 &  & True & & False &  & \multirow{2}*{\text{Semi-positive}}  \\ 
    2 &  & False & & True & &  \\ 
    3 &  & False & & False & & Negative  \\ 
\bottomrule
\end{tabular}}
\end{center}
\end{table}

\vspace{-0.5em}

\subsubsection{Multimodal Contrastive Learning}

We aim to reduce the multimodal prediction bias caused by treating modalities equally for each sample. This is done by learning to align unimodal representations towards the effective modalities sample by sample. We regulate each unimodal representation with the targets based on the multi-task-based multimodal learning framework. As shown in Figure \ref{fig03} c), we propose a new multimodal contrastive method to encourage modalities with both correct unimodal predictions to share a stronger correspondence. For those samples with both wrong predictions, we encourage their unimodal representations to be more different from each other to get more complementary multimodal representations. It helps to a higher possibility of correct multimodal prediction. For those samples with mutually exclusive predictions, we encourage these unimodal representations to learn from each other under the supervision of unimodal predictions by aligning the ineffective modality with the effective one.  

When considering two specific modalities $m_a$ and $m_b$ of $n$-th sample, we generate two unimodal hidden representations $r_a^n$ and $r_b^n$ from respective unimodal encoders. From the above unimodal predicting step, we also obtain the unimodal prediction results, $\hat{y}_a^n$ and $\hat{y}_b^n$. As the summarization in Table \ref{tab00}, we define the following positive pair, negative pair and semi-positive pair:

\textbf{Positive pair.} If both the paired unimodal predictions are correct, we define these  unimodal representation pairs are positive pairs, namely $\mathbb{P}$, where $\mathbb{P} = \{ n| \{ \hat{y}_a^n \equiv y^n \text{ and }  {\hat{y}_b^n \equiv y^n}\}_{n=1}^N \} $ in the mini-batch $\mathbb{B}$.

\textbf{Negative pair.} If both the paired unimodal predictions are wrong, we define these  unimodal representation pairs are negative pairs, namely $\mathbb{N}$, where $\mathbb{N} = \{ n | \{ {\hat{y}_a^n \neq y^n} \text{ and } {\hat{y}_b^n \neq y^n} \}_{n=1}^N \}$ in the mini-batch $\mathbb{B}$. 

\textbf{Semi-Positive pair.} If the predictions of the paired unimodal representations are mutually exclusive, one correct and another wrong, we define these unimodal representation pairs are semi-positive pairs, namely $\mathbb{S}$, where   $\mathbb{S} = \{ \{ n| \{ {\hat{y}_a^n \equiv y^n} \text{ and } {\hat{y}_b^n \neq y^n} {\}}_{n=1}^N \} \cup \{ n| {{\hat{y}}_a^n \neq y^n} \\ \text{ and }  {\hat{y}_b^n \equiv y^n} \}_{n=1}^N \} \}$  in the mini-batch.

We further propose the multimodal contrastive loss for two modalities as follows:
\begin{equation} \small
    \begin{aligned}
        \mathcal{L}_{b-mmc} =  -\log{\frac{\sum_{n\in{\mathbb{P, S}}}(\exp({\text{cos}(r_a^n,{r_b^n})/{\tau}})}{\sum_{n\in{\mathbb{B}}}(\exp({\text{cos}(r_a^n, {r_b^n})/{\tau}})}},
        \label{eq:eq07}
    \end{aligned}
\end{equation}
where $\text{cos}(r_a^n,r_b^n) = \frac{r_a^n\cdot{r_b^n}}{\lVert r_a^n\rVert \ast \lVert r_b^n\rVert}$ is the cosine similarity between paired unimodal representations $r_a^n$ and $r_b^n$ for sample $n$, ${\tau}$ is the temperature coefficient. The similarity of positive pairs and semi-positive pairs is optimized towards a higher value while the similarity of negative pairs is optimized towards a smaller value.  The difference between positive and semi-positive pairs is that the unimodal representations updated towards each other in positive pairs while only the unimodal representations of the wrong unimodal prediction updated towards the correct one in semi-positive pairs. We detach the modality feature with correct predictions from the computation graph when aligning with low-quality modality features for semi-positive pairs, which is inspired by GAN models \cite{arjovsky2017wasserstein, zhu2017unpaired} where the generator output is detached when updating the discriminator only,

Multimodal problems often encounter situations with more than two modalities. For more than two modalities, the multimodal contrastive loss for $M$ modalities ($M>2$) can be computed by:

\begin{equation} \small
    \begin{aligned}
        \mathcal{L}_{mmc} = \sum_{i=1}^{M}\sum_{j>i}^{M}\mathcal{L}_{b-mmc}(m_i, m_j),
        \label{eq:eq08}
    \end{aligned}
\end{equation}

\subsection{Fusion and Total Learning Objective}

\textbf{Multimodal prediction.} When fusing all unimodal representations with concatenation, we get the fused multimodal representations $r_c = r_1 \oplus r_2 \oplus ... \oplus r_m$. Similarly,  the multimodal predictive distribution is the output of the multimodal classifier with inputs of the fused representations. For the multimodal prediction task, the target is to minimize the multimodal prediction loss:

\begin{equation} \small
    \begin{aligned}
        \mathcal{L}_{multi} = -\sum_{k=1}^{K}{y}^{k}\log{p_k^k},
        \label{eq:eq05}
    \end{aligned}
\end{equation}
where $y^{k}$ is the $k$-th element category label and $[p_k^{1};p_k^{2};...;p_k^{K}] = \mathbf{p}_c(\mathbf{y}\, |\, \mathbf{r}_c)$ is the softmax output of multimodal classifier. 

\textbf{Total learning objective.}
The overall optimization objective for our proposed UniS-MMC is:
\begin{equation} \small
    \begin{aligned}
        \mathcal{L}_{UniS-MMC} = \mathcal{L}_{uni} + \mathcal{L}_{multi} + \lambda \mathcal{L}_{mmc},
        \label{eq:eq09}
    \end{aligned}
\end{equation}
where $\lambda$ is a loss coefficient for balancing the predicting loss and the multimodal contrastive loss.

\section{Experiments}
\subsection{Experimental Setup}
\textbf{Dataset and metric.} We evaluate our method on two publicly available image-text classification datasets UPMC-Food-101 and N24News. \textbf{UPMC-Food-101} \footnote{UPMC-Food-101:  https://visiir.isir.upmc.fr/} is a multimodal classification dataset that contains textual recipe descriptions and the corresponding images for 101 kinds of food. We get this dataset from their project website and split 5000 samples from the default training set as the validation set. \textbf{N24News} \footnote{N24News:  https://github.com/billywzh717/N24News} is an news classification dataset with four text types (\textit{Heading}, \textit{Caption}, \textit{Abstract} and \textit{Body}) and images. In order to supplement the long text data of the FOOD101 dataset, we choose the first three text sources from N24News in our work. We use classification accuracy (Acc) as evaluation metrics for UPMC-Food-101 and N24News. The detailed dataset information can be seen in Appendix \ref{appdix_1}.

\begin{table*}[t] \small
\caption{Comparison of multimodal classification performance on {\color{blue} \textbf{a)}} Food101 and {\color{blue} \textbf{b)}} N24News.}
\label{tab01}
\begin{center}
\setlength\tabcolsep{1.5pt}
\scalebox{0.9}{\begin{tabular}{lcccccc} 
    \toprule
    \multirow{2}*{{\color{blue} \textbf{a)}}  \textbf{Model}}  &  \multicolumn{2}{c}{\textbf{Fusion}} & &  \multicolumn{2}{c}{\textbf{Backbone}} & \multirow{2}*{\textbf{Acc}}  \\ 
    \cmidrule{2-3}  \cmidrule{5-6}
     & AGG  & ALI & & Image & Text &  \\
    \midrule
    MMBT & Early & \xmark & & ResNet-152 & BERT & $\text{92.1}_{\pm 0.1}$   \\
    HUSE & Early & \cmark & & Graph-RISE & BERT & $\text{92.3}$   \\
    ViLT & Early & \cmark & & ViT & BERT & $\text{92.0}$   \\
    CMA-CLIP & Early & \cmark & & ViT  & Transformer & $\text{93.1}$   \\
    ME & Early & \xmark & & DenseNet & BERT & $\text{94.6}$   \\
    \midrule
    AggMM & Early & \xmark & & ViT & BERT & $\text{93.7}_{\pm 0.2}$   \\
    UnSupMMC & Early & \cmark & & ViT & BERT & $\text{94.1}_{\pm 0.7}$   \\
    SupMMC & Early & \cmark & & ViT & BERT & $\text{94.2}_{\pm 0.2}$   \\
    \midrule
    \textbf{UniS-MMC} & Early & \cmark  & & ViT & BERT &  $\textbf{94.7}_{\pm 0.1}$   \\
    \bottomrule
\end{tabular}}
\quad
\scalebox{0.9}{\begin{tabular}{lccccccccccc}
    \toprule
    \multirow{2}*{{\color{blue} \textbf{b)}}  \textbf{Model}} &  \multicolumn{2}{c}{\textbf{Fusion}} & &  \multicolumn{2}{c}{\textbf{Backbone}} & & \multicolumn{3}{c}{\textbf{Multimodal}}  \\ 
    \cmidrule{2-3}  \cmidrule{5-6} \cmidrule{8-10}
    & AGG  & ALI & & Image & Text &  & Headline & Caption & Abstract \\
    \midrule
    N24News & Early & \xmark & & ViT & RoBERTa &  & $\text{79.41}$ & $\text{77.45}$ & $\text{83.33}$  \\
    \midrule
    AggMM & Early  &\xmark & & ViT & BERT &  & $\text{78.6}_{\pm 1.1}$ & $\text{76.8}_{\pm 0.2}$ & $\text{80.8}_{\pm 0.2}$ \\
    UnSupMMC & Early & \cmark & & ViT & BERT & & $\text{79.3}_{\pm 0.8}$ & $\text{76.9}_{\pm 0.3}$ & $\text{81.9}_{\pm 0.3}$ \\
    SupMMC & Early & \cmark & & ViT & BERT & & $\text{79.6}_{\pm 0.5}$ & $\text{77.3}_{\pm 0.2}$ & $\text{81.7}_{\pm 0.8}$ \\
    \text{UniS-MMC} & Early &\cmark & & ViT & BERT &  & $\textbf{80.2}_{\pm 0.1}$ & $\textbf{77.5}_{\pm 0.3}$ & $\textbf{83.2}_{\pm 0.4}$ \\
    \midrule
    AggMM & Early  &\xmark & & ViT & RoBERTa &  & $\text{78.9}_{\pm 0.3}$ & $\text{77.9}_{\pm 0.3}$ & $\text{83.5}_{\pm 0.2}$ \\
    UnSupMMC & Early  &\cmark & & ViT & RoBERTa &  & $\text{79.9}_{\pm 0.2}$ & $\text{78.0}_{\pm 0.1}$ & $\text{83.7}_{\pm 0.3}$ \\
    SupMMC & Early & \cmark & & ViT & RoBERTa & & $\text{79.9}_{\pm 0.4}$ & $\text{77.9}_{\pm 0.2}$ & $\text{84.0}_{\pm 0.2}$  \\
    \textbf{UniS-MMC} & Early & \cmark & & ViT & RoBERTa & & $\textbf{80.3}_{\pm 0.1}$ & $\textbf{78.1}_{\pm 0.2}$ & $\textbf{84.2}_{\pm 0.1}$  \\
    \bottomrule
\end{tabular}}
\end{center}
\end{table*}

\textbf{Implementation.} For the image-text dataset UPMC Food-101, we use pretrained BERT \cite{devlin-etal-2019-bert} as a text encoder and pretrained vision transformer (ViT) \cite{dosovitskiy2021an} as an image encoder. For N24News, we utilize two different pretrained language models, BERT and RoBERTa \citep{liu2019roberta} as text encoders and also the same vision transformer as an image encoder. All classifiers of these two image-text classification datasets are three fully-connected layers with a ReLU activation function.

The default reported results on image-text datasets are obtained with BERT-base (or RoBERTa-base) and ViT-base in this paper. The performance is presented with the average and standard deviation of three runs on Food101 and N24News. The codes is available on GitHub \footnote{https://github.com/Vincent-ZHQ/UniS-MMC}. The detailed settings of the hyper-parameter are summarized in Appendix \ref{appdix_2}.

\subsection{Baseline Models}
The used baseline models are as follows:
\begin{itemize}[leftmargin=*]
\setlength\itemsep{0em}
    \item  \textbf{MMBT} \cite{kiela2019supervised} jointly finetunes pretrained text and image encoders by projecting image embeddings to text token space on BERT-like architecture.
    \item  \textbf{HUSE} \cite{narayana2019huse} creates a joint representation space by learning the cross-modal representation with semantic information.
    \item  \textbf{ViLT} \cite{kim2021vilt, Liang_2022_CVPR} introduces a BERT-like multimodal transformer architecture on vision-and-language data. 
    \item  \textbf{CMA-CLIP} \cite{liu2021cma} finetunes the CLIP \cite{radford2021learning} with newly designed two types of cross-modality attention module. 
    \item  \textbf{ME} \cite{Liang_2022_CVPR} is the state-of-the-art method on Food101, which performs cross-modal feature transformation to leverage cross-modal information. 
    \item  \textbf{N24News} \cite{wang-etal-2022-n24news} train both the unimodal and multimodal predicting task to capture the modality-invariant representations. 
    \item  \textbf{AggMM} finetunes the pretrained text and image encoders and concatenates the unimodal representations for the multimodal recognition task. 
    \item  \textbf{SupMMC} and \textbf{UnSupMMC} finetune the pretrained text and image encoders and then utilize the supervised and unsupervised multimodal contrastive method to align unimodal representations before creating joint embeddings, respectively. 
\end{itemize}

\subsection{Performance Comparison}

\textbf{Final classification performance comparison.} The final image-text classification performance on Food101 and N24News is presented in Table \ref{tab01}. We have the following findings from the experimental results: (i) focusing on the implemented methods, contrastive-based methods with naive alignment could get an improvement over the implemented aggregation-based methods;  (ii) the implemented contrastive-based methods outperform many of the recent novel multimodal methods; (iii) the proposed UniS-MMC has a large improvement compared with both the implemented contrastive-based baseline models and the recent start-of-art multimodal methods on Food101 and produces the best results on every kind of text source on N24News with the same encoders.

\textbf{T-sne visualization comparison with baseline models.} We visualize the representation distribution of the proposed uni-modality supervised multimodal contrastive method and compare it with the naive aggregation-based method and the typical unsupervised and supervised contrastive method. 

\begin{figure}[h]
    \centering    \includegraphics[width=1\columnwidth]{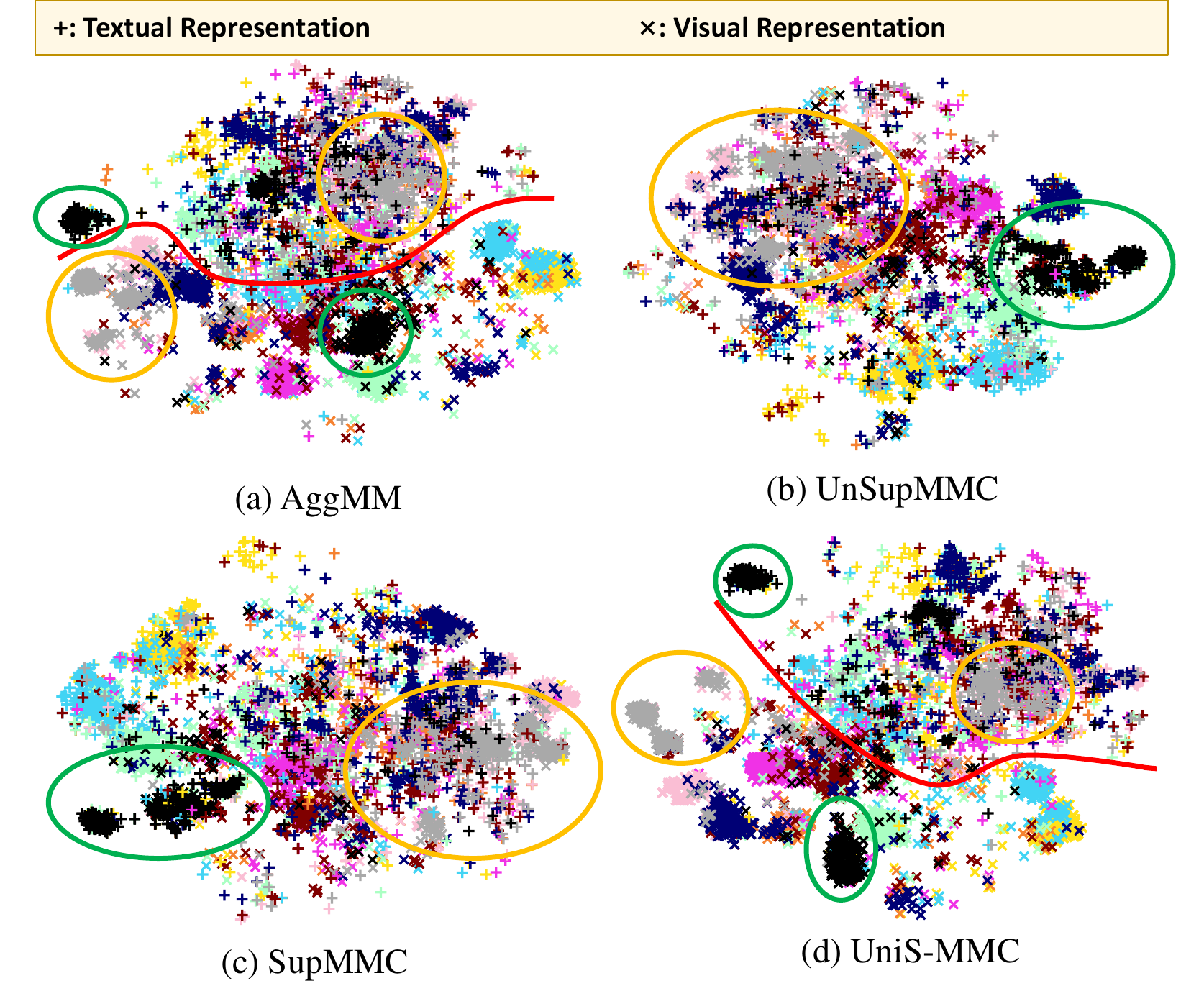}
    \caption{Unimodal representation distribution of the first 10 categories of the N24News test set across different methods: (a) aggregation-based method, (b) unsupervised multimodal method, (c) supervised contrastive method and (d) unimodality-supervised method.}
    \label{fig06}
\end{figure}

As shown in Figure \ref{fig06}, unimodal representations are summarized and mapped into the same feature space. 
The previous typical contrastive methods, such as unsupervised and supervised contrastive methods will mix up different unimodal representations from different categories when bringing the representation of different modalities that share the same semantics closer. For example, the representations of two modalities from the same category are clustered well in Figure \ref{fig06} (b) and (c) (\textcolor[rgb]{0.0, 0.66, 0.47}{green circle} and \textcolor[rgb]{1.0, 0.75, 0.0}{orange circle}). However, these contrastive-based methods can also bring two problems. One is that they map the unimodal embeddings into the same embedding space will lose the complementary information from different modalities. Another is that they heavily mix the representations from the specific class with other categories, such as the clusters (\textcolor[rgb]{1.0, 0.75, 0.0}{orange circle}). As a comparison, our proposed method preserves the complementary multimodal information by maintaining the two parts of the distribution from two modalities (\textcolor{red}{red line}) well (Figure \ref{fig06}  (d) from the aggregation-based methods (Figure \ref{fig06} (a)) in addition to a better cluster of unimodal representations. 

\begin{figure}[htb]
    \centering    \includegraphics[width=1\columnwidth]{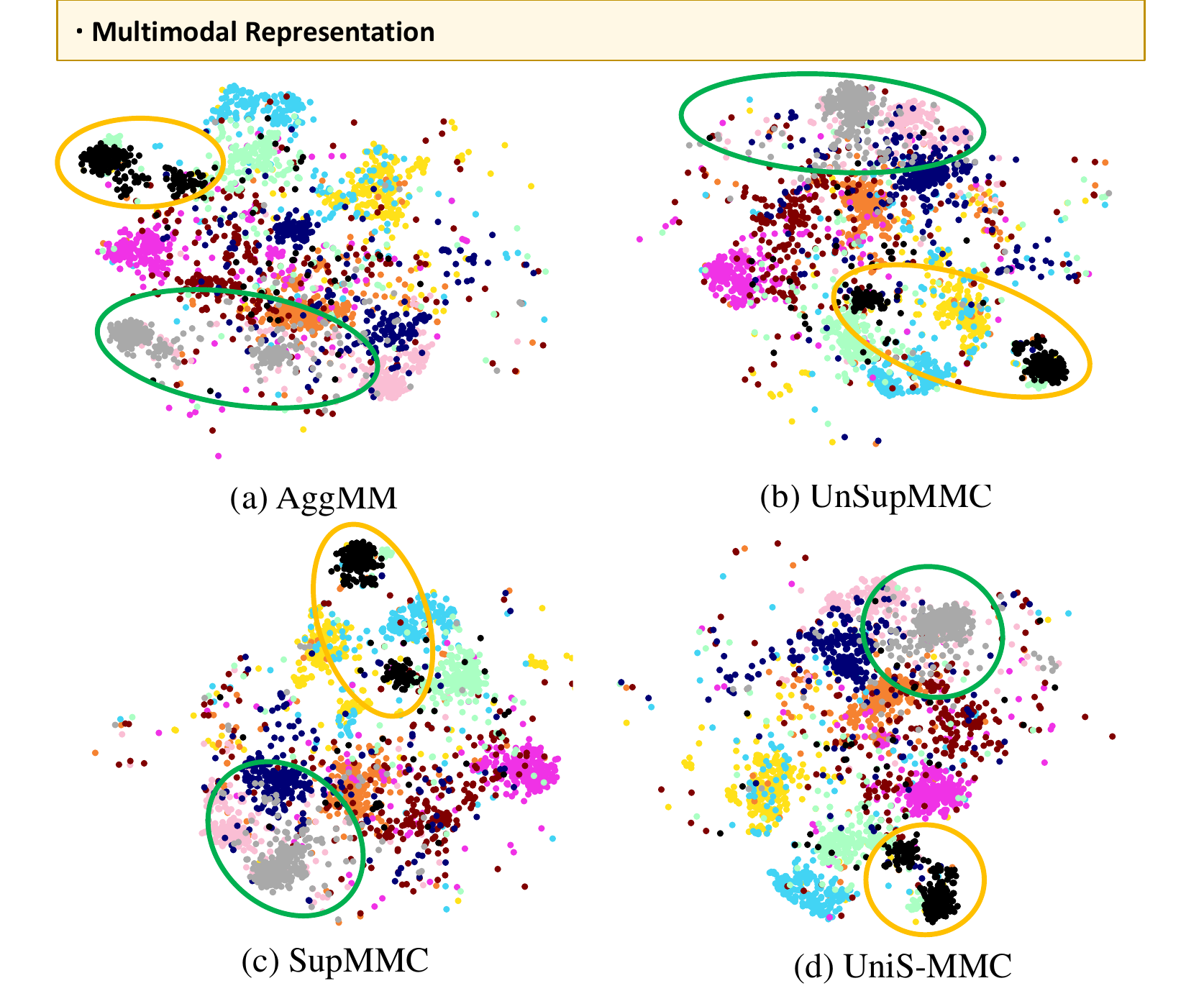}
    \caption{Multimodal representation distribution of the first 10 categories of the N24News test set across different methods: (a) aggregation-based method, (b) unsupervised multimodal method, (c) supervised contrastive method and (d) unimodality-supervised method.}
    \label{fig07}
\end{figure}

We further summarized the visualization of the final multimodal representation in Figure \ref{fig07}. Comparing Figure \ref{fig07} (a) and Figure \ref{fig07} (d), the proposed UniS-MMC can create better class clusters, such as the \textcolor[rgb]{0.0, 0.66, 0.47}{green circle}. Comparing Figure \ref{fig07} (b), (c) and (d) (\textcolor[rgb]{1.0, 0.75, 0.0}{orange circle}), the classification clusters are not separated by other classes in the proposed methods. It is different from the other two typical contrastive-based methods. Generally,  our proposed method not only helps the unimodal representation learning process and gets better sub-clusters for each modality but also improves the classification boundary of the final multimodal representation.

\subsection{Analysis}

\textbf{Classification with Different Combinations of Input Modalities.} We first perform an ablation study of classification on N24News with different input modalities. Table \ref{tab03} provides the classification performance of unimodal learning with image-only, text-only, and traditional multimodal learning with the concatenation of visual and textual features and our proposed UniS-MMC. The text modality is encoded with two different encoders, RoBERTa or BERT. By comparing the models with different language encoders, we find that the feature encoder can significantly affect the multimodal performance, and the RoBERTa-based model usually performs better than the BERT-based model. This is because the multimodal classification task is influenced by each learned unimodal representation. Besides, all the multimodal networks perform better than unimodal networks. It reflects that multiple modalities will help make accurate decisions. Moreover, our proposed UniS-MMC achieves $ 0.6\%$ to $2.4\%$ improvement over the aggregation-based baseline model with BERT and  $ 0.3\%$ to $1.4\%$ improvement with RoBERTa.

\begin{table*}[h]
\caption{Comparison to unimodal learning and the baseline model on N24News.}
\vspace{-0.5em}
\setlength\tabcolsep{5pt}
\label{tab03}
\begin{center}
\scalebox{0.83}{
\begin{tabular}{lccccccccccc}
    \toprule
    \multirow{2}{*}{\textbf{Dataset}} & \multirow{2}{*}{\textbf{Text}} & & \multirow{2}{*}{\textbf{Image-only}} & & \multicolumn{3}{c}{\textbf{BERT-based}} & & \multicolumn{3}{c}{\textbf{RoBERTa-based}}  \\
    \cmidrule{6-8}  \cmidrule{10-12}
    & & & & & Text-only & AggMM & \textbf{UniS-MMC} & & Text-only & AggMM  & \textbf{UniS-MMC} \\
    \midrule
    \multirow{3}{*}{\textbf{N24News}} & Headline & & \multirow{3}{*}{$\text{54.1}_{\pm 0.2}$} & & $\text{72.1}_{\pm 0.2}$ & $\text{78.6}_{\pm 1.1}$  & $\textbf{80.2}_{\pm 0.1}$ \color{blue} $\uparrow$ 1.6 & & $\text{71.8}_{\pm 0.2}$ & $\text{78.9}_{\pm 0.3}$ & $\textbf{80.3}_{\pm 0.1}$ \color{blue} $\uparrow$ 1.4\\
    & Caption & & & & $\text{72.7}_{\pm 0.3}$ & $\text{76.8}_{\pm 0.2}$ & $\textbf{77.5}_{\pm 0.3}$ \color{blue} $\uparrow$ 0.7 & & $\text{72.9}_{\pm 0.4}$ & $\text{77.9}_{\pm 0.3}$ & $\textbf{78.1}_{\pm 0.2}$ \color{blue} $\uparrow$ 0.3 \\
    & Abstract & & & & $\text{78.3}_{\pm 0.3}$ & $\text{80.8}_{\pm 0.2}$ & $\textbf{83.2}_{\pm 0.4}$ \color{blue} $\uparrow$ 2.4 & & $\text{79.7}_{\pm 0.2}$ & $\text{83.5}_{\pm 0.2}$ & $\textbf{84.2}_{\pm 0.1}$ \color{blue} $\uparrow$ 0.7 \\
    \bottomrule
\end{tabular}}
\end{center}
\end{table*}

\textbf{Ablation study on N24News.} We conduct the ablation study to analyze the contribution of the different components of the proposed UniS-MMC on N24News. AggMM is the baseline model of the aggregation-based method that combines the unimodal representation directly. The ablation works on three text source headline, caption and abstract with both BERT-based and RoBERTa-based models. Specifically, $L_{uni}$ is the introduced unimodal prediction task, $C_{Semi}$ and $C_{Neg}$ are semi-positive pair and negative pair setting, respectively.

Table \ref{tab04} presents the multimodal classification results of the above ablation stud with different participating components. $L_{uni}$ and the setting of $C_{Semi}$ align the unimodal representation towards the targets, with the former achieved by mapping different unimodal representations to the same target space and the latter achieved by feature distribution aligning. They can both provide a significant improvement over the baseline model. $C_{Neg}$ further improve the performance by getting a larger combination of multimodal representation with more complementary information for those samples that are difficult to classify.

\begin{table}[h] \small
\caption{Ablation study on N24News.}
\vspace{-1em}
\setlength\tabcolsep{1pt}
\label{tab04}
\begin{center}
\scalebox{0.86}{\begin{tabular}{lcccccccccc}
\toprule
    \multirow{2}*{\textbf{Method}} &  & \multicolumn{2}{c}{Headline} & &  \multicolumn{2}{c}{Caption} & &  \multicolumn{2}{c}{Abstract}  \\
    \cmidrule{3-4} \cmidrule{6-7}  \cmidrule{9-10}
    & & BERT & RoBERTa & & BERT & RoBERTa & & BERT & RoBERTa  \\
    \midrule
    AggMM &  & $\text{78.6}_{\pm 1.1}$ & $\text{78.9}_{\pm 0.3}$ & & $\text{76.8}_{\pm 0.2}$ & $\text{77.9}_{\pm 0.3}$ & & $\text{80.8}_{\pm 0.2}$ & $\text{83.5}_{\pm 0.2}$  \\ 
    + $L_{uni}$  & & $\text{79.4}_{\pm 0.4}$ & $\text{79.4}_{\pm 0.3}$  & & $\text{77.3}_{\pm 0.2}$ & $\text{77.9}_{\pm 0.1}$ &  & $\text{82.5}_{\pm 0.3}$ & $\text{84.1}_{\pm 0.2}$\\    
    + $C_{Semi}$  &  & $\text{80.1}_{\pm 0.1}$ & $\text{80.0}_{\pm 0.3}$ & & $\text{77.3}_{\pm 0.2}$ & $\text{78.0}_{\pm 0.3}$ &  & $\text{82.7}_{\pm 0.4}$ & $\text{84.2}_{\pm 0.3}$\\ 
    + $C_{Neg}$ & & $\textbf{80.2}_{\pm 0.1}$ &  $\textbf{80.3}_{\pm 0.1}$ & & $\textbf{77.5}_{\pm 0.3}$ & $\textbf{78.1}_{\pm 0.2}$ & & $\textbf{83.2}_{\pm 0.4}$ & $\textbf{84.2}_{\pm 0.1}$\\ 
\bottomrule
\end{tabular}}
\end{center}
\end{table}

\begin{figure}[!h]
    \centering    \includegraphics[width=1\columnwidth]{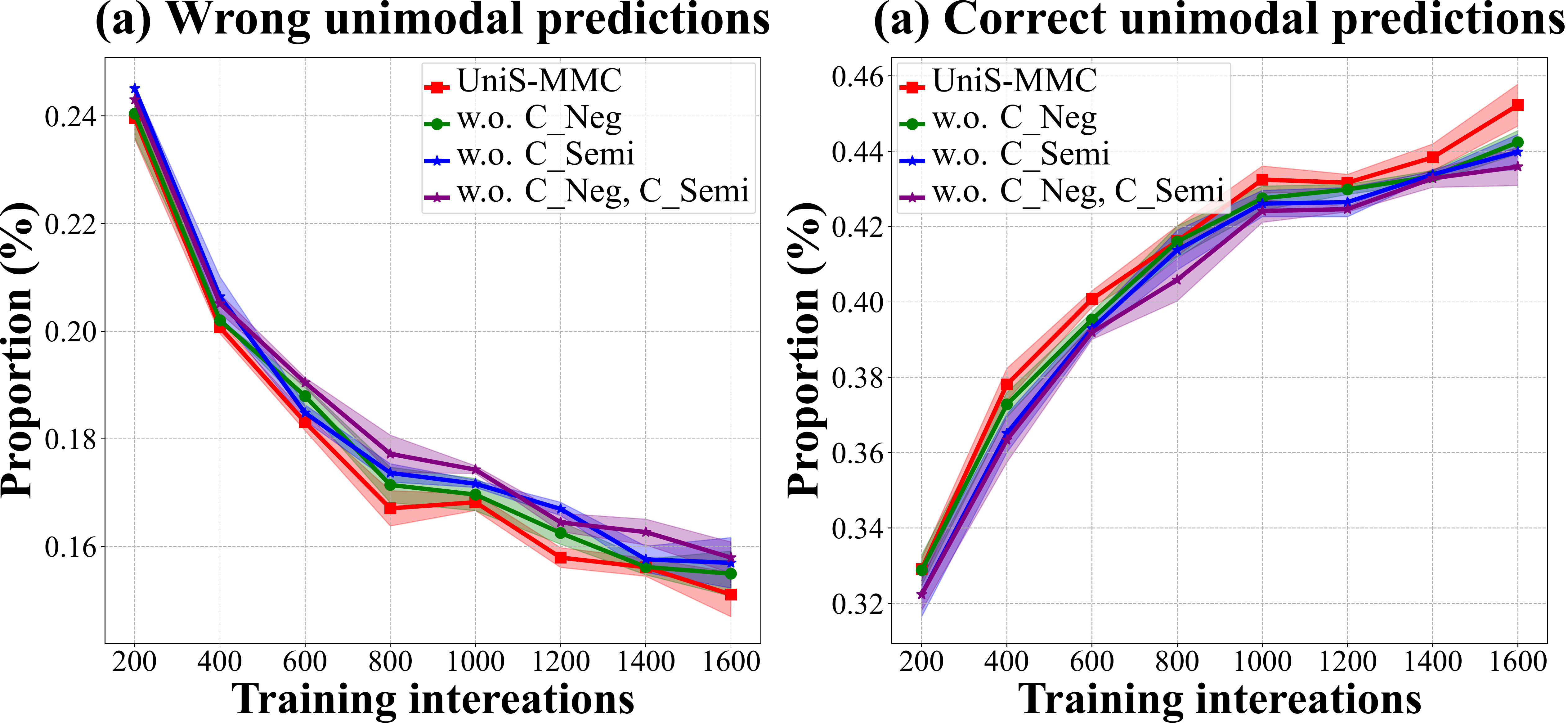}
    \caption{As the training progresses, the change of the proportion of both wrong (left), both correct (right) unimodal predictions of the validation set (N24News): the complete method (UniS-MMC), remove negative pair (w.o. C\_{Neg}), remove semi-positive pair (w.o. C\_{Semi}) and remove both  (w.o. C\_{Neg},C\_{Semi}).}
    \label{fig05}
\end{figure}

\textbf{Analysis on the learning process.} To further explore the role of our proposed UniS-MMC in aligning the unimodal representation towards the targets, we summarise the unimodal predicting results of the validation set during the training process in Figure \ref{fig05}. Ideally, different participating modalities for the same semantic should be very similar and give the same answer for the same sample. However, in practical problems, the unimodal predictions are not usually the same as the actual noise. In our proposed method, the proportion of both wrong unimodal predicting is higher and the proportion of both correct unimodal predicting is lower when removing our setting of semi-positive pair and negative pair. It means that UniS-MMC could align the unimodal representations for the targets better and get more trustworthy unimodal representations. 

\textbf{Analysis on the Final Multimodal Decision.}  Compared with the proposed UniS-MMC, MT-MML is the method that jointly trains the unimodal and multimodal predicting task, without applying the proposed multimodal contrastive loss. We summarize unimodal performance on MT-MML and UniS-MMC and present unimodal predictions in Figure \ref{fig05}. The unimodal prediction consistency here is represented by the consistency of the unimodal prediction for each sample. When focusing on the classification details of each modality pair, we find that the proposed UniS-MMC gives a larger proportion of samples with both correct predictions and a smaller proportion of samples with both wrong decisions and opposite unimodal decisions compared with MT-MML.

\begin{figure}[h]
    \begin{center} \includegraphics[width=1\columnwidth]{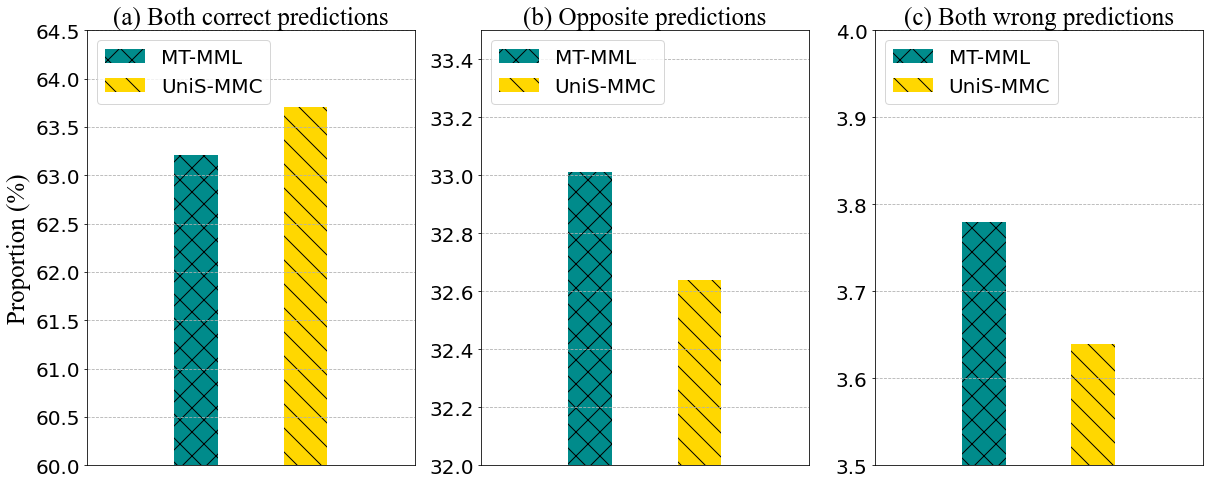}
    \end{center}
    \caption{Consistency comparison of unimodal prediction between MT-MML and the UniS-MMC.}
    \label{fig08}
\end{figure}

\vspace{-1em}

\section{Conclusion}
In this work, we propose the Unimodality-Supversied Multimodal Contrastive (UNniS-MMC), a novel method for multimodal fusion to reduce the multimodal decision bias caused by inconsistent unimodal information. Based on the introduced multi-task-based multimodal learning framework, we capture the task-related unimodal representations and evaluate their potential influence on the final decision with the unimodal predictions. Then we contrastively align the unimodal representation towards the relatively reliable modality under the weak supervision of unimodal predictions. This novel contrastive-based alignment method helps to capture more trustworthy multimodal representations.  The experiments on four public multimodal classification datasets demonstrate the effectiveness of our proposed method.

\section*{Limitations}
Unlike the traditional multimodal contrastive loss focusing more on building the direct link between paired modalities, our proposed UniS-MMC aims to leverage inter-modality relationships and potential effectiveness among modalities to create more trustworthy and complementary multimodal representations. It means that UniS-MMC is not applied to all multimodal problems. It can achieve competitive performance in tasks that rely on the quantity of the joint representation, such as the multimodal classification task.  It is not suitable for tasks that rely purely on correspondence between modalities, such as the cross-modal retrieval task.

\section*{Acknowledgements}
The computational work for this article was partially performed on resources of the National Supercomputing Centre, Singapore (https://www.nscc.sg)

\bibliography{acl2023}
\bibliographystyle{acl_natbib}

\appendix

\section{Appendix}
\label{sec:appendix}

\subsection{Datasets Usage Instructions} \label{appdix_1}
To make a fair comparison with the previous works, we adopt the following default setting of the split method, as shown in Table \ref{tab:appdix_1}. Since the UPMC-Food101 dataset does not provide the validation set, we split 5000 samples out of the training set and use them as the validation set.

\begin{table}[htb] \small
\caption{Datasets information and the split results}
\label{tab:appdix_1}
\begin{center}
\scalebox{0.86}{\begin{tabular}{c|ccccc}
\toprule
    Dataset  & Modalities & \#Category & \#Train & \#Valid & \#Test \\
    \midrule
    UPMC-Food-101 & image, text & 101 & 60085 & 5000 & 21683 \\
    N24News & image, text & 24 & 48988 & 6123 & 6124 \\
\bottomrule
\end{tabular}}
\end{center}
\end{table}

\subsection{Experimental Settings} \label{appdix_2}
The model is trained on NVIDIA V100-SXM2-16GB and NVIDIA A100-PCIE-40GB. The corresponding Pytorch version, CUDA version and CUDNN version are 1.8.0, 11.1 and 8005 respectively. We utilize Adam as the optimizer and use ReduceLROnPlateau to update the learning rate.  We use Adam \cite{DBLP:journals/corr/KingmaB14} as the model optimizer. The temperature coefficient for contrastive learning is set as 0.07 and the loss coefficient in this paper is set as 0.1 to keep loss values in the same order of magnitude. The code is attached and will be available on GitHub. Some key settings of the model implementation are listed as followings: 

\begin{table}[htb] \small
\caption{Detailed setting of the hyper-parameter for UPMC-Food-101, BRCA and ROSMAP}
\label{tab:appdix_2}
\begin{center}
\setlength\tabcolsep{15pt}
\scalebox{0.86}{\begin{tabular}{c|cccc}
\toprule
Item  & UPMC-Food-101 & N24News   \\
\midrule
    Batch gradient & 128 & 128  \\
    Batch size & 32 & 32 \\
    Learning rate (m) & 2e-5 & 1e-4  \\
    Dropout (m) & 0 & 0  \\
    Weight decay  & 1e-4 & 1e-4  \\
\bottomrule
\end{tabular}}
\end{center}
\end{table}

\subsection{Learning with a Single Modality} \label{appdix_3}
We show the unimodal classification results from different unimodal backbones on text-image datasets in the following Table \ref{tab:appdix_3}.

\begin{table}[htb] \small
\caption{Unimodal classification performance with different backbones on Food101 and N24News.}
\setlength\tabcolsep{10pt}
\label{tab:appdix_3}
\begin{center}
\scalebox{0.86}{\begin{tabular}{ccccccccccc}
\toprule
    \textbf{Source} & \textbf{Backbone} & 
    \textbf{Food101} & & \textbf{N24News}  \\
    \midrule
    \textbf{Image} & ViT & $\text{73.1}_{\pm 0.2}$ & &  $\text{54.1}_{\pm 0.2}$ & \\
    \midrule
    \textbf{Text} & BERT & $\text{86.8}_{\pm 0.2}$ & & - \\
    \midrule
    \multirow{2}*{\textit{Heading}}  & BERT & - & &$\text{72.1}_{\pm 0.2}$ \\
    & RoBERTa & - & & $\text{71.8}_{\pm 0.2}$ & \\
    \midrule
    \multirow{2}*{\textit{Caption}} &  BERT & - & & $\text{72.7}_{\pm 0.3}$ \\
    & RoBERTa & - &  &  $\text{72.9}_{\pm 0.4}$ \\
    \midrule
    \multirow{2}*{\textit{Abstract}} & BERT & -  &  & $\text{78.3}_{\pm 0.3}$ \\
    & RoBERTa & - &  & $\text{79.7}_{\pm 0.2}$ \\
\bottomrule
\end{tabular}}
\end{center}
\end{table}

\end{document}